# A modular framework for automated evaluation of procedural content generation in serious games with deep reinforcement learning agents

Eleftherios Kalafatis, Konstantinos Mitsis, Konstantia Zarkogianni, *Member, IEEE,* Maria Athanasiou, *Member, IEEE,* and Konstantina Nikita, *Fellow, IEEE*

***Abstract*— Serious Games (SGs) are nowadays shifting focus to include procedural content generation (PCG) in the development process as a means of offering personalized and enhanced player experience. However, the development of a framework to assess the impact of PCG techniques when integrated into SGs remains particularly challenging. This study proposes a methodology for automated evaluation of PCG integration in SGs, incorporating deep reinforcement learning (DRL) game testing agents. To validate the proposed framework, a previously introduced SG featuring card game mechanics and incorporating three different versions of PCG for non-player character (NPC) creation, has been deployed. Version 1 features random NPC creation while versions 2 and 3 utilize a genetic algorithm approach. These versions are used to test the impact of different dynamic SG environments on the proposed framework's agents. The obtained results highlight the superiority of the DRL game testing agents trained on Versions 2 and 3 over those trained on Version 1 in terms of win rate (i.e. number of wins per played games) and training time. More specifically, within the execution of a test emulating regular gameplay, both Versions 2 and 3 peaked at a 97% win rate and achieved statistically significant higher (p=0.009) win rates compared to those achieved in Version 1 that peaked at 94%. Overall results advocate towards the proposed framework's capability to produce meaningful data for the evaluation of procedurally generated content in SGs.**

## I. Introduction

PROCEDURAL Content Generation (PCG) techniques are considered valuable tools in the field of game development, offering the means to algorithmically create game content with limited or indirect user input [1]. By automating the process of generating game elements such as levels, maps, mechanics, textures, narrative, and more, PCG facilitates game developers in overcoming challenges related to manual content creation. Furthermore, experience-driven PCG has gained popularity, as a way to create content that can be tailored to each player's preferences, enhancing thus the overall game experience [2]. This approach utilizes modern techniques such as evolutionary algorithms and neural networks to model the player's state and needs [3].

Besides entertainment games, PCG is rapidly becoming a useful tool in serious games (SGs) as well. SGs are defined as digital games with a main purpose outside pure entertainment [4]. The incorporation of PCG techniques in SGs enables the rapid and efficient generation of content, facilitating the creation of engaging and personalized game interventions adaptable to the players' needs [5], [6]. One of the earliest examples of adaptive PCG for SGs is SIREN [7], a conflict resolution game that uses procedurally generated conflicts and narrative to tailor the game to specific player needs. Similarly, a more recent approach [8] tackles narrative generation, with emphasis on the scenario's conclusion, as a way to improve personalization on SGs by exploring interactive digital narratives, player experience modelling and experience-driven PCG.

The expected benefits of the procedurally generated game environments trade off against the great challenge of evaluating them [9]. Game testing (GT), employing both manual and automated approaches, is a crucial phase in the game development lifecycle, and involves the systematic evaluation of game mechanics, graphics, audio, performance, and other essential aspects [10]. GT can be employed to assess procedurally generated game content as well, ensuring its quality, functionality, and overall game experience. However, procedurally generated game content challenges manual GT techniques, in the sense that it increases the effort required to fully control and playtest a sometimes-infinite space of possible configurations. This is especially true in the case of SGs, where in addition to traditional testing goals, it should also evaluate the effectiveness of the intervention. One possible solution in such dynamically changing game settings is the use of automated GT techniques empowered by deep reinforcement learning (DRL) [11].

Employment of deep learning techniques in games has

E. Kalafatis, is with the School of Electrical & Computer Engineering, National Technical University of Athens, Athens 15772, Greece (e-mail: leftkal@biosim.ntua.gr)

K. Mitsis, is with the School of Electrical & Computer Engineering, National Technical University of Athens, Athens 15772, Greece (e-mail: kmhtshs@biosim.ntua.gr)

K. Zarkogianni, is with the Department of Advanced Computing Sciences, Faculty of Science and Engineering, Maastricht University, Maastricht 6200 MD, Netherlands (e-mail: konstantia.zarkogianni@maastrichtuniversity.nl)

M. Athanasiou, is with the School of Electrical & Computer Engineering, National Technical University of Athens, Athens 15772, Greece (e-mail: mathanasiou@biosim.ntua.gr)

K. S. Nikita, is with the School of Electrical & Computer Engineering, National Technical University of Athens, Athens 15772, Greece (e-mail: knikita@ece.ntua.gr)

recently been investigated, with applications in many settings, from PCG [12] to GT [13]. Reinforcement Learning, diverging from other ML approaches, empowers autonomous agents through trial-and-error interaction to acquire optimal learning policies in dynamic environments [14]. Building upon the foundations of deep learning and reinforcement learning, DRL leverages deep neural networks to tackle complex decision-making problems [15]. In general, the employment of deep learning architectures enables agents to learn high-dimensional representations of states and actions, facilitating more effective policy optimization [16]. This not only enhances the agent's ability to perceive and understand complex environments but also enables the exploration of more nuanced strategies and policies [17], [18]. DRL agents consist of deep neural networks, which endorse them with the capacity to mimic human playthroughs and in some cases even exhibit human-like attention [19]. In GT settings, they have demonstrated the ability to emulate human players' playstyles and skill levels, providing a more accurate representation of real-world gameplay scenarios [20], [21]. As a result, their ability to learn and successfully complete a vast variety of tasks makes them a desirable asset in the game development industry.

In SGs incorporating PCG, these human-like agents are exposed to differentiating content, oftentimes aiming to guide the learning process by varying the SG's difficulty. Of rather high importance is this dynamic difficulty adjustment's (DDA's) potential to improve the SG's learning outcome, as observed in a recent review paper [22]. Through these rich and dynamic interactions, metrics relevant to the nature of procedurally generated content (e.g. DDA, generated game elements) can be obtained to help improve the explainability of the PCG technique's choices and evaluate its effectiveness. These advantages motivated the exploitation of DRL agents' capacity to playtest procedurally generated environments in games [11], [23]. In one recent approach, DRL agents have been used to imitate player's capacity and to continuously raise the difficulty in video games [24].

However, to the best of our knowledge, there is no reported attempt to develop a scalable, flexible and modular evaluation framework for assessing the effectiveness of SG incorporating PCG. The present study investigates the use of DRL GT agents towards the development of a methodology for the evaluation of PCG driven SGs. DRL agents are employed for automated GT and the impact of their exposure to different versions of PCG, featuring random and a genetic algorithm (GA) based content generation, is explored. The conducted experiment hypothesis is that agents exposed to the GA PCG versions, designed to provide content gradually and relative to the player's level, are expected to achieve better metrics compared to those exposed to random content generation. A conceptual framework addressing the interactions between the SG and the agents along with the extraction and visualization of metrics to aid human evaluation of the GT process has been designed.

## II. Related Work

Several studies focusing on processes that facilitate testing and evaluation of procedurally generated content have been reported. Overall, according to [2], there exist three main methodological testing and evaluation approaches: (i) the direct, (ii) the simulation, and (iii) the interactive evaluation functions. Direct evaluation is achieved by establishing a set of measurable metrics or properties. One of the earliest direct approaches [25] performs a comparison of different PCG level generators for the Mario AI framework. The study introduces a number of metrics such as leniency, linearity, density, pattern density, pattern variation, compression distance, that set a baseline for automatic evaluation of the generated platformer levels. Another study [26] investigates further on this type of level generators, by exploring methods for testing, visualizing, and selecting the generated content, based on its distribution, robustness and expressive range. Direct evaluation can also be applied to analyze optimization problems in PCG. The employment of a generator for Super Mario Bros levels coupled with different tools to assess its capabilities, such as diagonal walks, the estimation of structural high-level properties and problem similarity measures, addresses the challenges of defining appropriate representations for game content and efficient evaluation functions, considering several factors such as the vast search space and the subjectivity inherent in human perception [27]. Direct evaluation approaches are consistent, but often difficult or even impossible to design and implement in large game spaces. This difficulty arises due to the complexity and inherent creative nature of generated game content.

Simulation-based and interactive evaluation of procedurally generated content is performed through a combination of data collected during game playing, either from artificial players (simulation-based) or human players (interactive), and expert evaluation. An example of simulation-based evaluation can be found in an online combat game called "Wuji". More specifically, a methodology involving exhaustive exploration of game states for automatic GT is introduced in a simulation environment, with the aim to identify logical bugs, gaming balance bugs, and user experience issues, shedding light on the evaluation process's practical challenges [23]. However, a review paper on simulation-based GT identifies a lack of research on the intrinsic motivations, such as competence, curiosity, autonomy and relatedness, of the autonomous agents [28]. Later papers try to tackle this problem using various methods such as MCTS agents with evolved heuristics that act as intrinsic motivation [29], or RL designed to emulated human-like play-styles [30] with success. On the other hand, a study employing interactive evaluation involved educational games and recruited 150 students to evaluate the accuracy of employing support vector machines and GA for PCG [31]. Another example of interactive evaluation is a study of procedurally generated levels of match-three games, visual analysis performed by experts and players is conducted to pre-screen the generated content and identify and evaluate patterns that may not be computationally measurable. Interactive evaluation has also been applied to identify properties characterizing the different game levels such as symmetry or patterns appearing in the training [32]. Comparing between simulation and interactive driven PCG evaluation approaches, the volume of necessary data to produce meaningful insight often makes the employment of human testers costly, especially in early development stages. Hence, the use of systematic

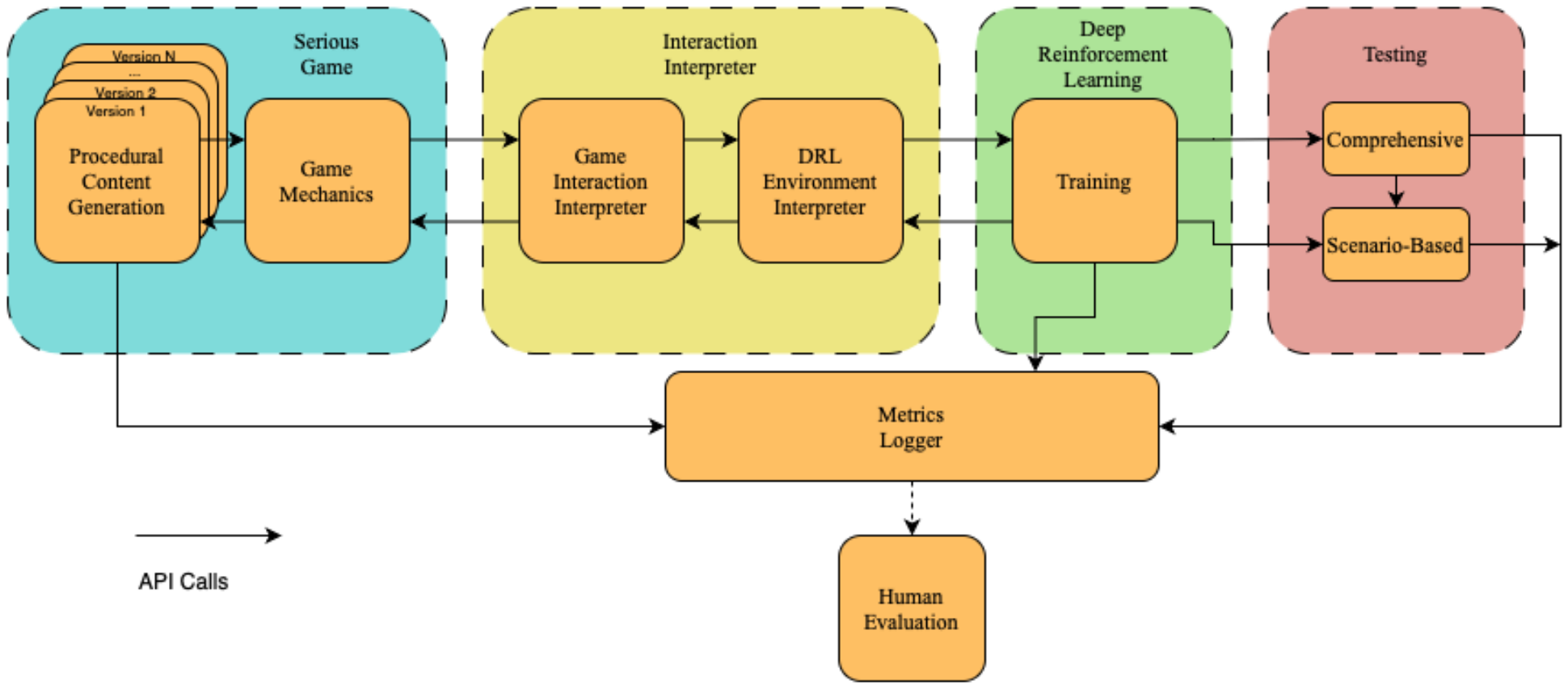

**Fig. 1.** The Evaluation Framework contains four spaces; Serious Game, Interaction Interpreter, Deep Reinforcement Learning and Testing. The Metrics Logger collects information from the Serious Game Deep Reinforcement Learning and Testing spaces and parses it for human evaluation.

frameworks for automated generation of testing data is an active field of research.

Within this context, a few frameworks have been proposed aiming to provide a systematic and generalized approach for PCG evaluation. The GVGAI framework [33], provides a programmatical interface for creating agents able to play any game written in VGDL, that is a specialty language created for describing games. A review on agents developed based on this framework [34], reveals that GVGAI facilitates the design of AI-assisted games along with automated game testing and debugging. This framework can also be applied for creating a testing pipeline that uses a team of agents implementing the MAP-Elites algorithm for automated game testing. Each one of the agents follows a different directive, such as winning, exploring, being curious, killing and collecting [35], [36]. Another promising framework that deals with automated GT is PathOS [37]. This framework has been developed as a tool for the Unity game engine, to enable the development of the agents as well as their easy integration in the game development phase with minimal extra effort. Lastly, a more general framework has been developed, that enables the utilization of different computer vision algorithms to aid GT. It features two entities, a game bot and a state checker and is available as an open-source plugin that can be adjusted for different game engines [38]. No particular frameworks addressing the specific challenges of incorporating PCG is SGs have been identified.

## III. Conceptual Framework

A conceptual framework for the evaluation of PCG in SGs by employing DRL GT agents is presented (Fig. 1). The framework relies on the generation of GT data by training and evaluating DRL agents on different versions of PCG. These versions might feature distinct PCG techniques or differentiations of the same. Comparison between DRL agent performance yields insight regarding each version's effectiveness and allows for PCG technique selection and parameter tuning. Such a pipeline falls in line with simulation evaluation techniques. As opposed to traditional DL approaches [13], it leverages the unsupervised nature of DRL to create GT data to eliminate human input at least until the evaluation. The framework comprises four spaces that consist of several modules which communicate with each other through API interfaces.

The "Serious Game" space encapsulates the game logic, along with the graphic user interface. It communicates with the rest of the framework by implementing functions which handle game actions and game states. SGs are represented by Game Description Language (GDL) [39], a declarative logic programming language. PCG techniques are also implemented in their native form within this space, ensuring that both real-time and predetermined PCG can be evaluated.

In the "Interaction Interpreter" space, the game mechanics are translated and expressed in a DRL-specific format. The Game Interaction Interpreter (GII) expresses the characteristics of the SG in a quantifiable manner. It interfaces to the SG through its API and delivers three functions to the rest of the framework; "perform_action" that takes as input a vector of numbers and translates it into a valid game action, "get_state" that returns a vectorial representation of the game state and "get_representations" that returns two vectors representing the action and the state space, respectively. The DRL Environment Interpreter (DEI) is responsible for translating the GII output to the DRL agents. It communicates with the GII through the GII functions and has knowledge of the DRL architecture. It handles the SG observations, a variable that conveys the game state to the agent, the rewards for the agent, and the end conditions of the game.

The "Deep Reinforcement Learning" space incorporates the DRL GT agent training process. During training, DRL GT agents interact with the game and build knowledge upon the SG's learning objectives. It takes into consideration the game observations and produces Serious Game Actions (SGAs) that are communicated to the DEI in order for the latter to change the game environment. SGAs are also employed as a time measurement unit by the framework to signify DRL GT agent training completion.

Finally, GT data produced by multiple trainings are fed to the "Testing" space, where two processes occur, the

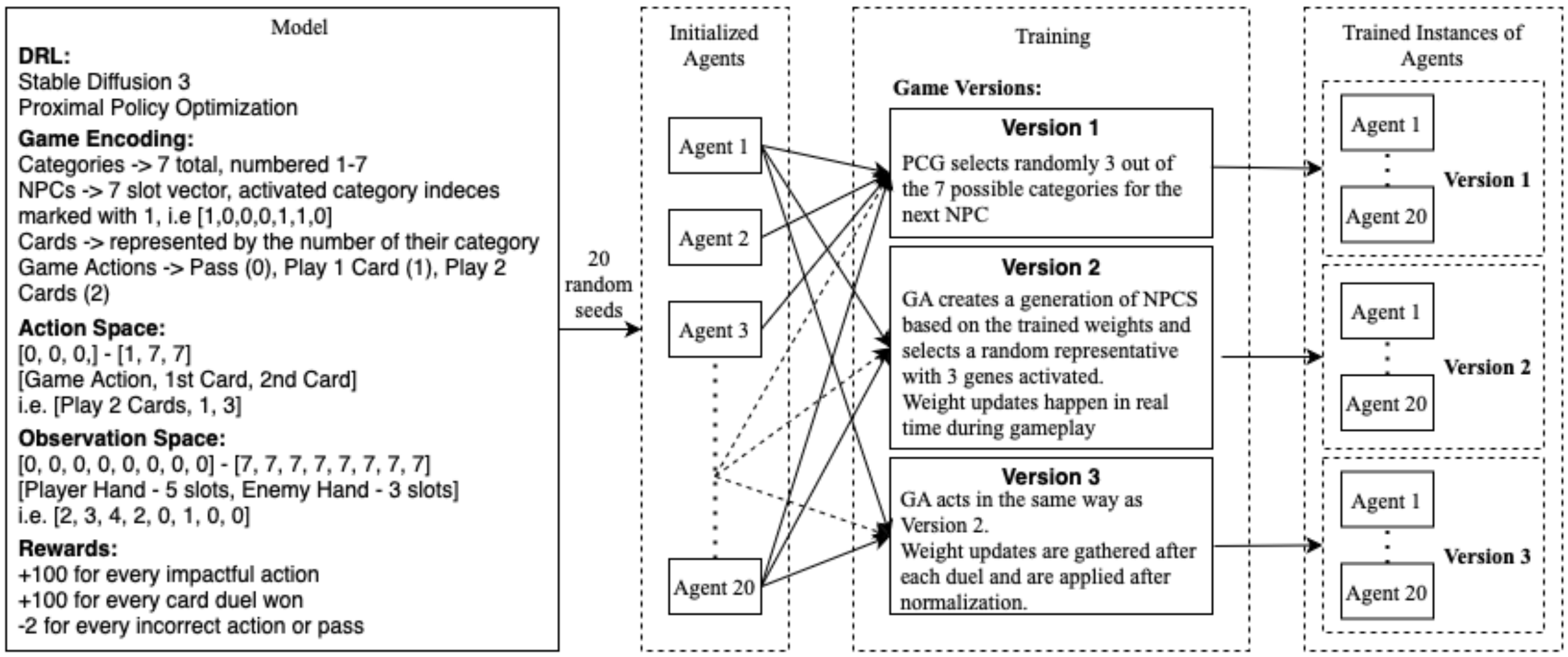

**Fig. 2.** Training Process Pipeline; the DRL Model (Stable Diffusion 3, PPO) is initialized with 20 random seeds to create the agents. The agents are then trained against three SG versions and checkpoints are saved throughout training. This results in a set of trained instances for each agent and each SG version.

Comprehensive Test and the Scenario-Based Test. The Comprehensive Test aims to expose DRL GT agents to an adequate volume of all possible generated SG content, characterized by high variability, to cater for high dimensional PCG techniques. The Scenario-Based Test places trained agents against playthroughs produced through PCG to realistically simulate actual play. This way, the performance comparison of GT agents trained by different versions of PCG is assessed extensively. Metrics produced during both test processes, GT training and PCG are gathered by the Metrics Logger and employed to provide insight regarding PCG performance.

The proposed framework is flexible, easily accommodating major changes throughout its entire pipeline. For instance, modifications in DRL model affect only the DEI part, while the deployment of a new SG would only affect the GII part. This modularity is achieved through the use of the suitable API interfaces capable of acting as means of communication and standardization among the different parts. As a result, the proposed framework can be rapidly deployed in different SGs featuring a variety of PCG techniques and provide data-driven insight regarding the quality of procedurally generated content without the use of human testers.

## IV. Experiment

For validation purposes, the proposed evaluation framework has been employed to compare the GT data produced by agents exposed on a SG incorporating different versions of PCG.

### *A. Serious Game Space*

The employed SG, "Wake Up for the Future" [40], aims to raise awareness about Obstructive Sleep Apnea (OSA). The SG features card game mechanics and an open world enabling the player to interact with non-player characters (NPCs). The player's task is to engage in debates simulated by card duels against NPCs with undiagnosed cases of OSA. Each NPC is associated with a unique profile enriched with information regarding daily habits, medical history and knowledge relating to OSA. All this OSA related information is split into categories, denoted as attributes (e.g. medication, hypertension, sleep position), to be handled by the SG's systems and mechanics. Based on these attributes, game cards are available to the player to be used in card duels, as a means to contradict NPC cards containing erroneous arguments about OSA. These NPC cards also correspond to attributes and are selected in each duel based on the NPC's profile. The player wins a duel if they successfully convince the NPC by contradicting all their arguments. If they run out of cards or a set number of turns passes without achieving this goal, they lose the duel.

The SG incorporates PCG to produce the NPC profiles by determining attributes in terms of amount and type. Three versions of the SG have been developed incorporating different methods for the generation of the NPC profiles for the experiment's purposes (Fig. 2). SG version 1 produces NPCs by selecting their attributes randomly. This version does not incorporate PCG, therefore it can be considered as the baseline for comparison. SG versions 2 and 3 employ a GA based approach for PCG, which selects NPC attributes dynamically [40]. As a result, one key difference between versions 2 and 3 against version 1, is the frequency of specific attribute selection. In a previous study [40], human testers experienced higher sense of competence and lower negative experience when interacting with "Wake Up for the Future" versions similar to 2 and 3 in comparison to version 1. The employment of these different SG versions enables the assessment of the impact of different PCG parameters in the training of DRL GT agents.

The GA based PCG generates NPCs by defining their attributes based on the player's interaction with the SG. To this end, genes are used to indicate the presence of attributes in a generated NPC. These genes receive binary values: "1" if a specific attribute is present in the NPC profile, and "0" if not. These genes are combined into a vector to form chromosomes, that represent a complete NPC profile. As described in [40], the GA's weight update functions receive data from the player's interaction with the NPC's attributes and adjust the gene weights. The initial generation of chromosomes is created randomly. All SG versions employed in the experiment feature

NPCs that possess 3 out of 7 available attributes. Following each debate, all chromosomes are evaluated through the fitness functions and 5 of them which score the best and are deemed the fittest from the previous generation are selected as parents to produce the next generation of 20 NPCs as offspring. The fitness functions receive as input the gene weights and promote game content either to challenge the player or to ease their experience and facilitate learning. To this end, the gene weights are trained from player interaction with the OSA related attributes and the player's performance. By design, every GA's generation after the first few converges to chromosomes with attributes that the PCG evaluates as suitable for the player. To create the content for each debate, the GA picks one of the 20 NPCs that were generated after the last debate randomly for the player to face [40].

The evaluation of the fittest NPCs relies on the weights of their activated genes. The weights undergo adjustments during gameplay based on fitness functions. The fitness functions monitor the interaction of the player with the SG and implement a set of rules. When the player is winning, for every NPC card not destroyed, the corresponding weight is increased; for every player card played correctly, the associated weight is decreased; for every player card played incorrectly, the respective weight is increased and for every player card left unplayed, the corresponding weight is increased. When the player is losing, for every NPC card not destroyed, the corresponding weight is decreased and for every player card left unplayed, the corresponding weight is increased. After calculating the weights, the fittest NPCs, which have the largest sum of gene weights, are paired up at random crossover points, and each pair generates two offspring. The number of fittest NPCs selected ensures that the population remains constant across generations. Each NPC has a 5% probability of undergoing a mutation within the resulting generation, which entails changing genes to their complementary value. This mutation mechanism introduces a degree of diversity and variability within the population.

TABLE I
DEEP REINFORCEMENT LEARNING ENVIRONMENT INTERPRETER INTERFACE

| Function | Description |
|---|---|
| init | initializes gymnasium environment and game |
| step | transmits the agent's action to the GII and returns the observations to the agent, along with the calculated scores |
| reset | resets states after a session is finished and prepares the variables the next session |

The difference between versions 2 and 3 (Fig. 2) lies in the update of the gene weights. In version 2 the GA leaves the update unmonitored, allowing for extreme updates, directly depicting the evaluation of the player's interaction with the game. On the other hand, version 3 tries to control this update, by gathering the individual updates after each duel and normalizing the mean value and standard deviation before applying them. The employment of these two versions attempts to provide insight regarding the proposed framework's capacity to discern between different implementation of the same PCG technique.

### *B. Interaction Interpreter Space*

The employed SG, initially developed in Unity, is adapted to allow for communication with the interpreters and integration with the workflow of the framework. To this end, the portion of the game that features PCG, is expressed in GDL and implemented using the pygdl Python library. The reasoner is then able to answer queries about the game state, responding with "true" or "false", depending on whether the query holds or not. A state machine is responsible for progressing the game state based on the player's actions. The state machine offers two functions, called "move" and "score", which have the responsibility to advance the state and return the score of the game respectively. To facilitate integration with the rest of the framework's components, the GA based PCG technique is also deployed in Python.

The GII takes the set of functions available from pygdl and standardizes the interaction with DEI, by offering the functions "perform_action", "get_state" and "get_representations". The first function, "perform_action" uses the "move" function of the state machine, together with specific knowledge about the SG architecture to advance the game state. The second, "get_state", queries the state machine to create a vectorized representation of the SG state. Lastly, "get_representations", is used for initialization purposes and to communicate the action and state space to the DEI. In the presented conceptual framework, the DEI is implemented using the gymnasium library [41]. This library provides a way to streamline the interaction of the agent with the rest of the framework by implementing 3 functions (TABLE I).

### *C. Deep Reinforcement Learning Space*

The employment of the DRL agents is based on the Stable-Baselines3 (SB3) framework [42] that features robustness and flexibility in complex DRL policies. These traits are of paramount importance to navigate the procedurally generated SG environments required to produced adequate testing data. SB3, an open-source framework, addresses this challenge by implementing seven commonly employed model-free DRL algorithms [43]. SB3 is compatible with gymnasium and uses the standard "learn" and "predict" functions for learning and providing predictions.

**Algorithm 1**: Agent Initialization

```
x := 20                          # number of agents to create
for i ← 1 to x do
|  # create instance of version 1
|  env := make_env(random)
|  # initialize agent for the environment
|  agent := PPO(MlpPolicy, env, n_steps=128,
|                 batch_size=4, vf_coef=0.5,
|                 clip_range=0.2)
|  # save agent to designated path
|  agent.save(path)
end
```

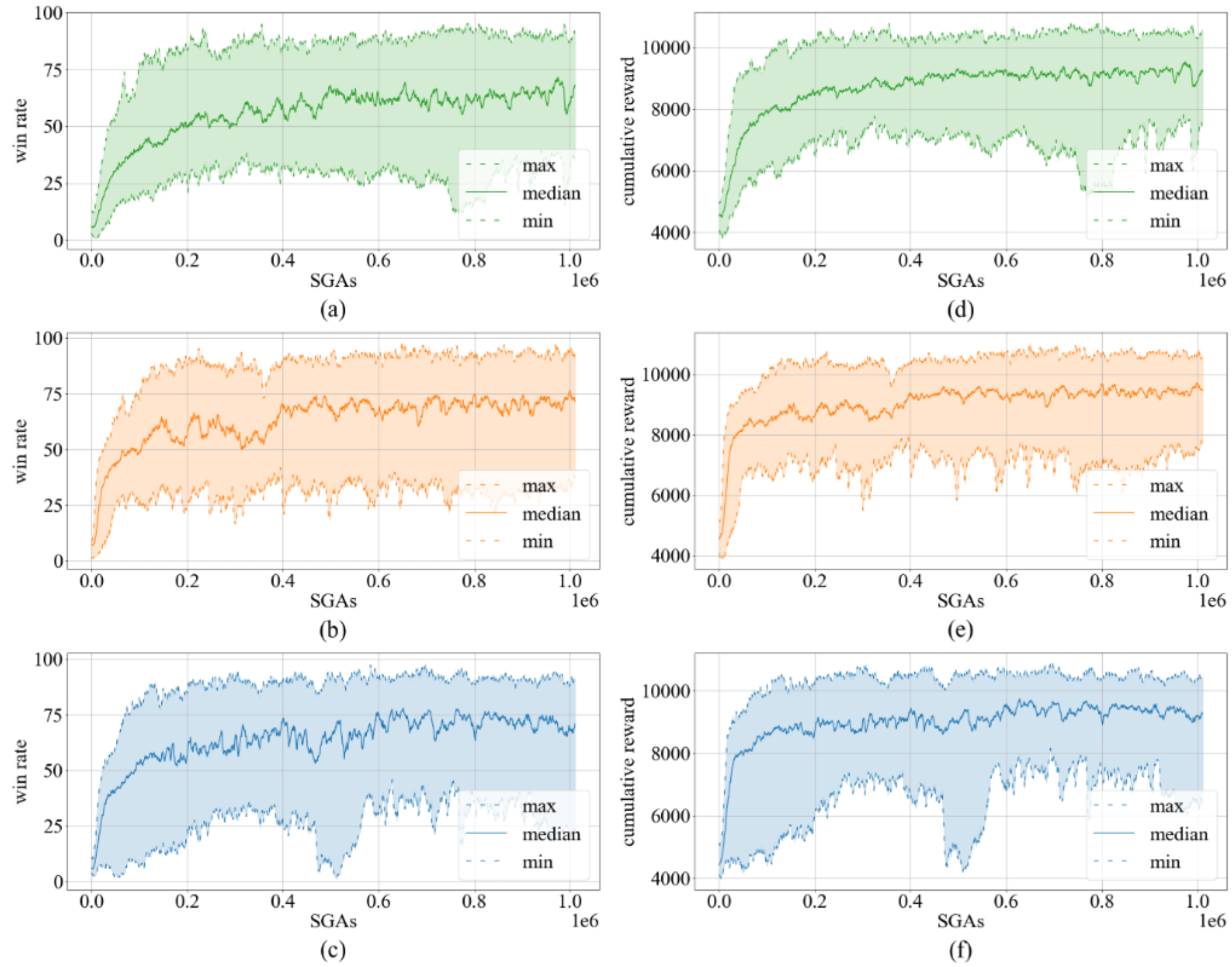

**Fig. 3.** The maximum, median and minimum win rate achieved by agents for (a) Version 1 (b) Version 2 (c) Version 3 and cumulative reward gathered by agents for (d) Version 1 (e) Version 2 (f) Version 3 calculated every 500 SGAs.

In the present study, the DRL algorithm is Proximal Policy Optimization (PPO). PPO algorithms are designed to learn optimal policies for decision-making in complex environments, such as game playing and robot control. One of the key features of PPO algorithms is their ability to update the policy more smoothly than other DRL algorithms. PPO algorithms also use an optimization objective that is simpler and easier to optimize than other DRL algorithms, further improving their stability and efficiency. Additionally, a recent finding suggests that the task performance of PPO agents and their correspondence to human attention are highly correlated [19], making them desirable for tasks producing SG testing data. However, it should be noted that PPO has not been specifically designed for GT applications.

In order to check the effectiveness of PPO's certain default hyperparameters (n_steps, batch_size, vf coef, clip_range [16]) an initial screening for the DRL training performance is conducted utilizing SG version 1. Based on the obtained results, a grid search is performed in small batches to further optimize the hyperparameters with correspondence to learning speed and performance. The agent initialization, along with the parameters used is presented in Algorithm 1, with implementation notes in Fig. 2. To communicate the SG's state and actions through the DEI, two vectors are used, taking values from the observation and action space respectively. After each SGA the agent receives a reward, calculated through trial-and-error, of +100 for every impactful action (as denoted by the GII) and for every duel won and a penalty of -2 for every other action (Fig. 2.). The slightly negative value of penalty further highlights the missed positive reward and nudges the agent towards impactful actions, without overly punishing strategies that include non-impactful moves. The impact of actions is evaluated by set rules during GII's development (e.g. the action destroys a NPC card).

## D. *Testing Space*

For the experiment's purposes, to cater for adequate power for the statistical analysis, through detailed search, a set of twenty (20) agents has been created by means of applying random seeds as depicted in Fig. 2. Each of these agents is trained in all three SG versions. Trial and error tests revealed that one million (1,000,000) SGAs are adequate for the completion of each training, since more SGAs do not lead to higher training performance. For every agent, a set of trained instances is collected during training every ten thousand (10,000) SGAs. A series of metrics, namely agent win rate (i.e. number of wins per played games), cumulative reward, number of games played and the frequency of appearance of each NPC attribute is gathered for every agent during training every five hundred (500) SGAs by the Metrics Logger.

For the Comprehensive Test, all possible NPC attribute combinations that include 3 out of the possible 7 attributes are generated manually. The order of appearance of NPC attributes, i.e. the specific permutation of the chosen attributes, affects agents' performance and is chosen by the SG. Due to this limitation, each trained agent is decided to play against each combination of NPC attributes five times, which means that with 55.5% probability it will face at least 3 out of the 6 different permutations, as a tradeoff to ensure an adequate representation of the possible opponents.

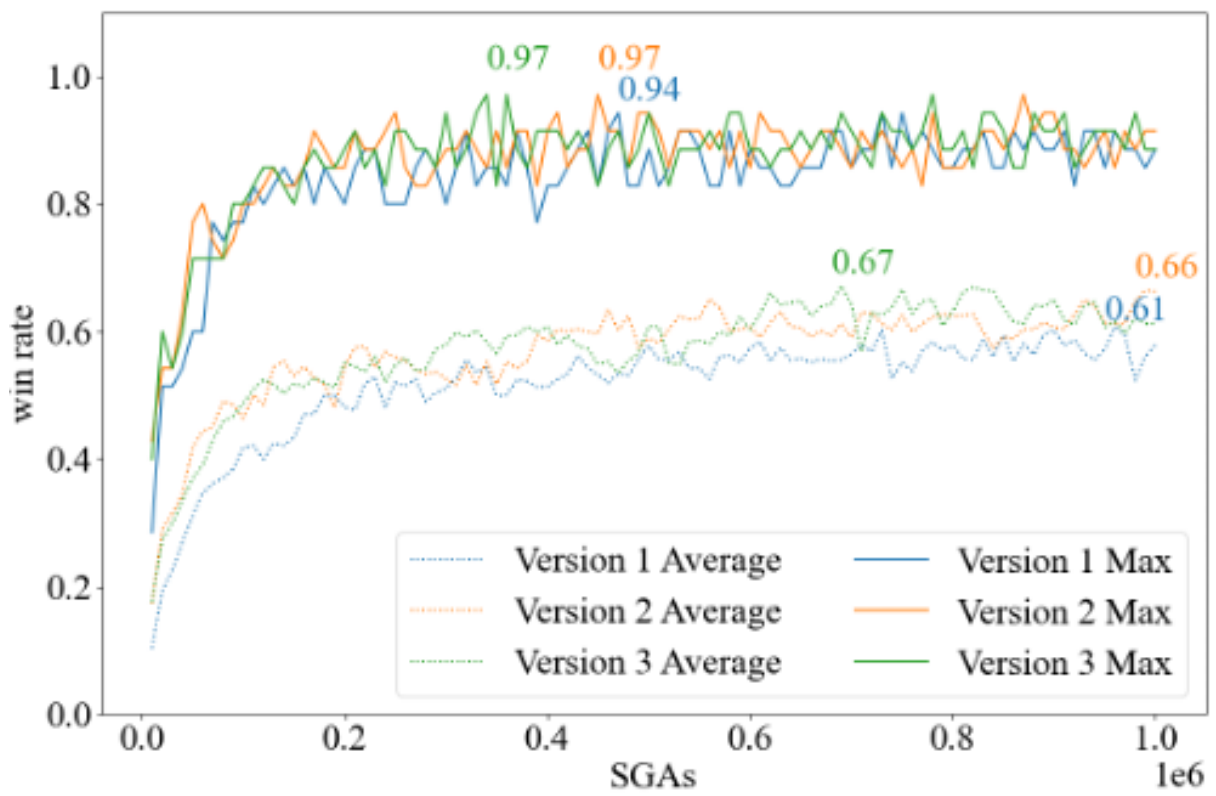

**Fig. 4.** Comprehensive Test. Dashed lines show the average win rate and regular lines show the max win rate of the agent instances trained for the respective number of SGAs.

The Scenario-Based test pits the best agent trained instances from all versions against NPCs generated randomly (version 1) and through GA based PCG (version 2). To assure a balanced comparison between agents trained in all SG versions, a selection of the best-performing trained instances, based on win rate achieved during the comprehensive test, is employed. In particular, the best trained instance of each agent trained on each version is picked. This results in 60 different agent instances, pitted against 1,000 NPCs generated by version 1 and 1,000 NPCs generated by version 2. Their win rate is gathered by the Metrics Logger. Consequently, according to these win rates, outlier agents are detected based on the interquartile range (IQR) method, which removes any data point 1.5 IQR points below the first quartile and 1.5 IQR points above the third quartile. To ensure fair comparison these outlier agents are removed (6 agents against randomly created NPCs and 4 against GA based PCG created NPCs), along with agents resulting from the same random seed.

## V. RESULTS

### A. Deep Reinforcement Learning Training

Results from DRL agent training for 1 million SGAs on each version are presented in Table II. The statistical significance threshold for all tests is set at 0.05. The total number of games completed by agents in each SG version during training differs slightly due to the variability of game duration and its dependence on agent winning efficiency. However, no statistically significant differences are observed. Kruskal-Wallis performed on total wins resulted in a p-value equal to 0.08, indicating no statistical significance.

In Fig. 3 the win rate for each version per 500 SGAs is presented. Dashed lines depict the minimum and maximum win rates, whereas the continuous line shows the median win rate across all agents. Median values were preferred over average values, to accommodate for the impact of outlier agent performance. It is evident that agents in all three versions achieved median win rates above 60%. Versions 2 and 3 display median values above 60% in most SGAs after four hundred thousand SGAs of training, as opposed to version 1.

Versions 1 and 3 both demonstrate policy collapse behavior, where an agent's win rate after some training deteriorates in comparison to its previous performance, and for an extensive period of SGAs stayed that way.

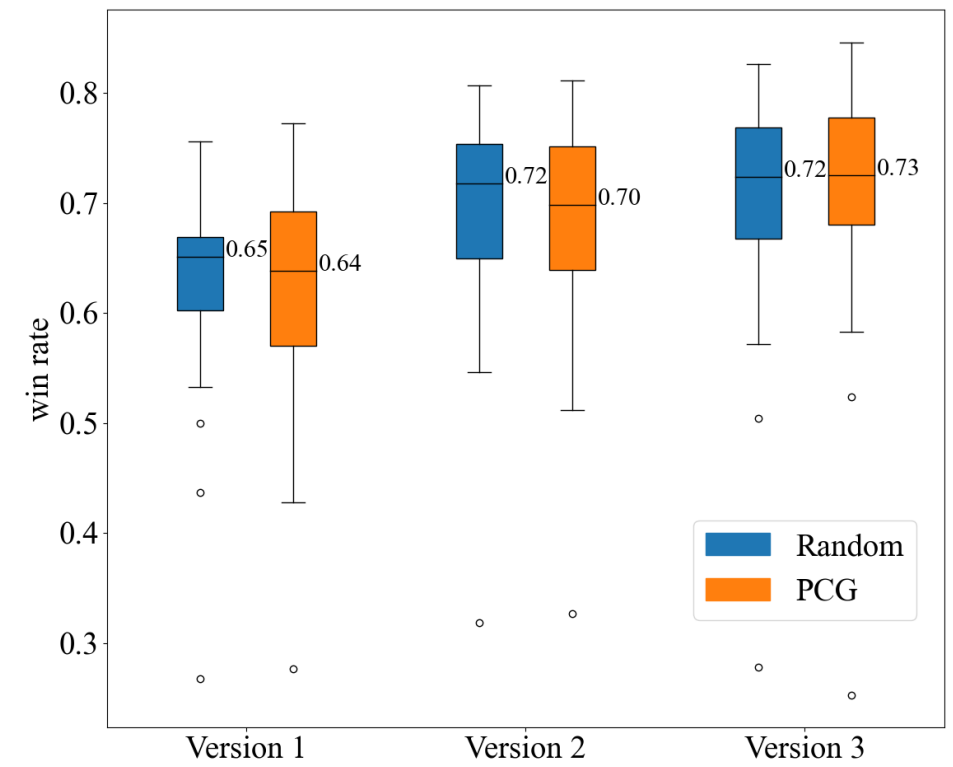

**Fig. 5.** Scenario-Based Test. Boxplot of the win rates of versions 1, 2 and 3 validated on randomly generated NPCs(blue) and NPCs generated by version 2 (orange)

This could happen when an agent becomes too specialized in countering a specific type of opponent, and fails to adapt to changes in its environment or opponent strategies, leading to performance degradation. Version 1 doesn't have any form of adaptation, and as such any form of collapsing policy has to run its course until it eventually starts learning again. Version 3, having the ability to differentiate the content based on the agent's interaction with the SG, should have been able to counteract such behaviors, but apparently the normalized and modest gene weight updates make it less flexible in that aspect than version 2.

Cumulative rewards diagrams demonstrate similar results (Fig. 3). Versions 2 and 3 achieve a better median cumulative reward both in terms of SGAs required and overall performance. In both figures no difference was presented regarding the max lines. In all three versions they behave similarly and stabilize around the same threshold. This could be a direct effect of the reward function, that rewards a win with the same amount as any other favorable move during gameplay, resulting in the negation of the small differences that emerge in the win rate.

TABLE II
TRAINING METADATA

| Version | Total Games | Total Wins | Avg. Reward per Game | Avg. Win rate |
|---|---|---|---|---|
| 1 | 4,395,586 | 2,263,205 | 79.50 ±3.81 | 0.51 ±0.09 |
| 2 | 4,398,076 | 2,491,408 | 82.15 ±2.59 | 0.57 ±0.06 |
| 3 | 4,374,053 | 2,480,172 | 81.76 ±2.85 | 0.57 ±0.07 |

### B. Comprehensive and Scenario-Based Test

Results from the Comprehensive test are depicted in Fig. 4. in terms of average and max win rate of the 20 agents trained on each version. It can be seen that the max win rates differ slightly between versions 2 and 3, both achieving a max win rate of 97%. In contrast, version 1 achieved a max win rate of 94% (Fig. 4). The average win rates depicted in Fig. 4 provide further insight regarding the agents' performance among versions. Specifically, version 2 and 3 both display superior performance over version 1 in almost all trained instances, as expected from the initial experiment hypothesis. They both achieve a win rate above 50% faster during training, and peak average win rate more than 5% higher when compared to version 1.

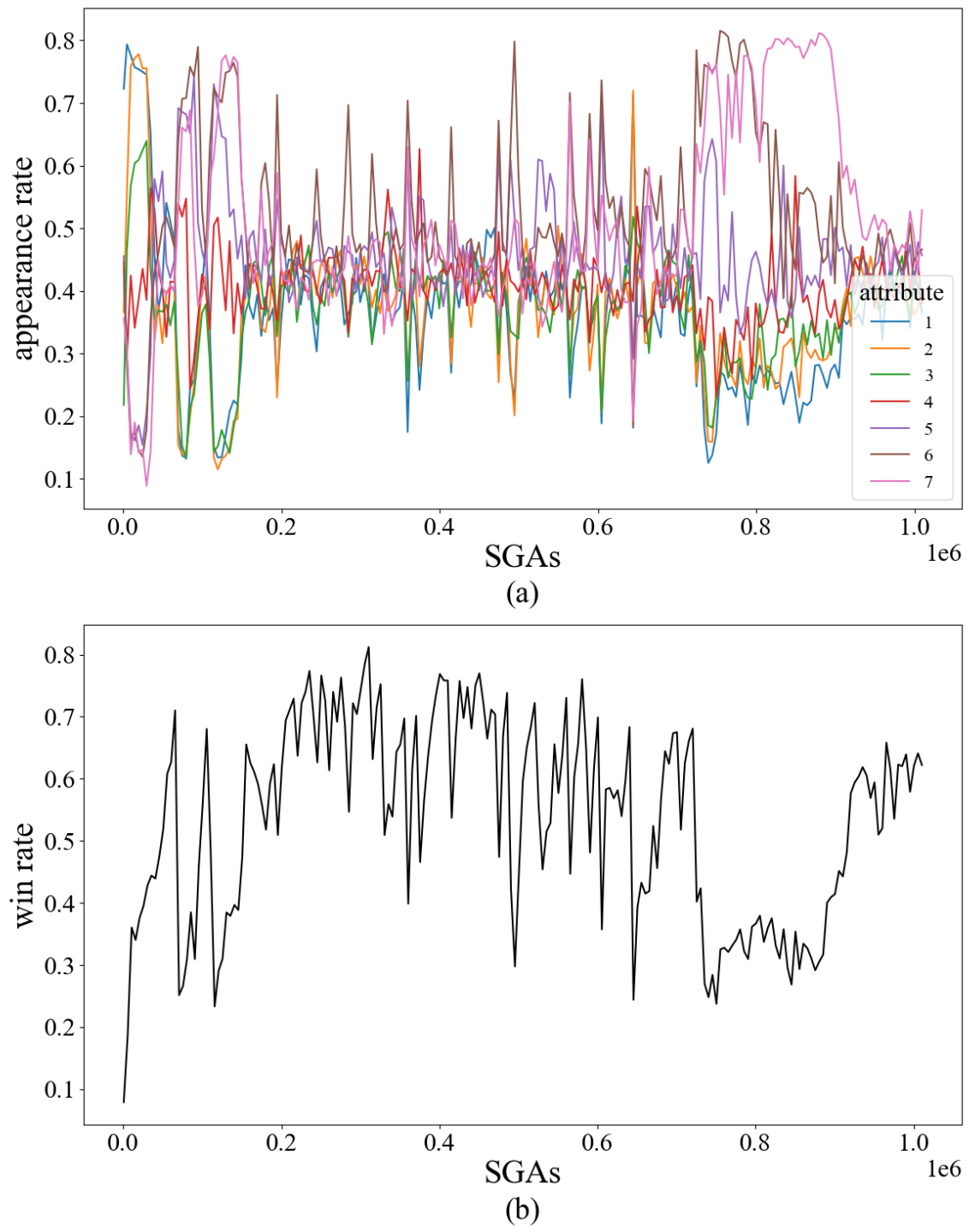

**Fig. 6.** PCG Performance – Version 2 (a) Appearance rate of attributes for agent 0 (b) Win rate of agent 0 over SGAs of training.

Results from the Scenario-Based test are presented in Fig. 5. Agents trained on versions 2 or 3 achieve better median win rates than the ones trained on version 1, when pitted against NPCs generated both randomly and through PCG, further validating the initial experiment hypothesis. For the statistical analysis, non-parametric tests were used, due to the Shapiro-Wilk test indicating non-normal distribution among the data. Kruskal-Wallis test reveals statistical significance on the win rates achieved by agents trained on different versions (p=0.007 for random NPCs, p=0.008 for PCG generated NPCs).

Mann-Whitney U tests reveal that this difference is particularly evident between agents trained on versions 1 and 3 (p=0.002 for random NPCs, p=0.015 for PCG generated NPCs). Similar inferences are obtained from applying Mann-Whitney U test between versions 1 and 2 (p=0.029 for random NPCs, p=0.040 for PCG generated NPCs). The performance of agents trained on the same SG version did not reveal statistically significant differences between random and PCG generated NPCs. This is expected after the completion of the training phase, as agents should display a stable performance.

### C. *PCG performance*

Fig. 6a presents an example of PCG attribute selection for one agent interacting with SG version 2 during training. Fig. 6b displays this agent's monitored win rate during training. Though this interpretation attempt of the inner workings of the GA, it is evident that changes in the selected attributes by the PCG correspond to changes in the agent's win rate. After each major change in the prevalent attributes, the agent's win rate drops. Similarly, when the agent's win rate starts rising, version 2 starts the process of rotating attributes. Similar results were identified in version 3.

Fig. 7 shows the median appearance rate of the attributes selected by version 2 across all agents during training. After the first 100k SGAs the rate of appearance for all attributes stabilizes around 42%.

This is an indication that the PCG is not driven into content saturation during training but retains versatility in attribute selection across multiple GT agents, as the theoretical probability of an attribute being selected in combinations of 3 out of 7 attributes without repetition equals to 42.9%.

## VI. Discussion

Win rates achieved in the first 200k SGAs of training indicate that DRL agents exposed to versions 2 (Fig. 1b) and 3 (Fig. 1c) achieve faster training compared to those exposed to version 1 (Fig. 1a). This observation may be related to versions' 2 and 3 capabilities to differentiate SG content when confronted with a winning agent, effectively adjusting game difficulty in the process. This DDA appears to benefit PPO's learning capacity and expedites GT agent training. This comes also in line with the decline of negative experience reported by human players facing a similar PCG technique in [40].

Similar observations have been reported on the effect of PCG driven DDA on human players [44]. This similarity could serve as an indication towards the potential of the proposed framework to produce meaningful insight regarding PCG in SGs. On the other hand, limited differences are observed between version 2 and version 3 during training. Agents trained in version 3 played a total of 24,023 less duels than those trained in version 2 (Table 1), hence a larger average number of SGAs was needed to achieve victory in version 3. Taking into consideration that version 3 agents achieved slightly better win rates as well (Fig. 1c), this could be interpreted as version 3 providing a higher degree of overall difficulty while maintaining smooth difficulty adjustment. This observation is consistent with the design direction applied in version 3, where fitness function weight updates are regulated to limit abrupt changes in rate of attribute appearance.

Based on the obtained results from both tests (Comprehensive and Scenario-Based), versions 2 and 3 achieve superior median win rate compared to version 1 (Fig. 4, 5). This is more evident in the Scenario-Based test, where Kruskal-Wallis and Mann-Whitney U results indicate statistically significant differences between version 1 and version 2- and 3-win rates. This observation is a strong indication towards the success of the GA based PCG to facilitate the DRL agents' learning process. When comparing between versions 2 and 3, maximum win rates don't display significant differences, especially during the exhaustive validation (Fig. 5).

Another finding obtained by the Scenario-Based test relates to the lack of statistically significant differences between the performance achieved when the same trained agent instance is pitted against opponents generated both randomly and through PCG (Fig. 5). This is an indication of the GA's ability to increase overall competence of the agents trained on all versions, as the content provided through PCG during testing doesn't affect the apparent win rate that these agents experience. This is possibly achieved by continuous and dynamic adjustment of the SG difficulty through PCG and comes in line with the results from a previous study with "Wake Up for the Future" with human testers [40] where GA version of the game scored higher in terms of sense of competence.

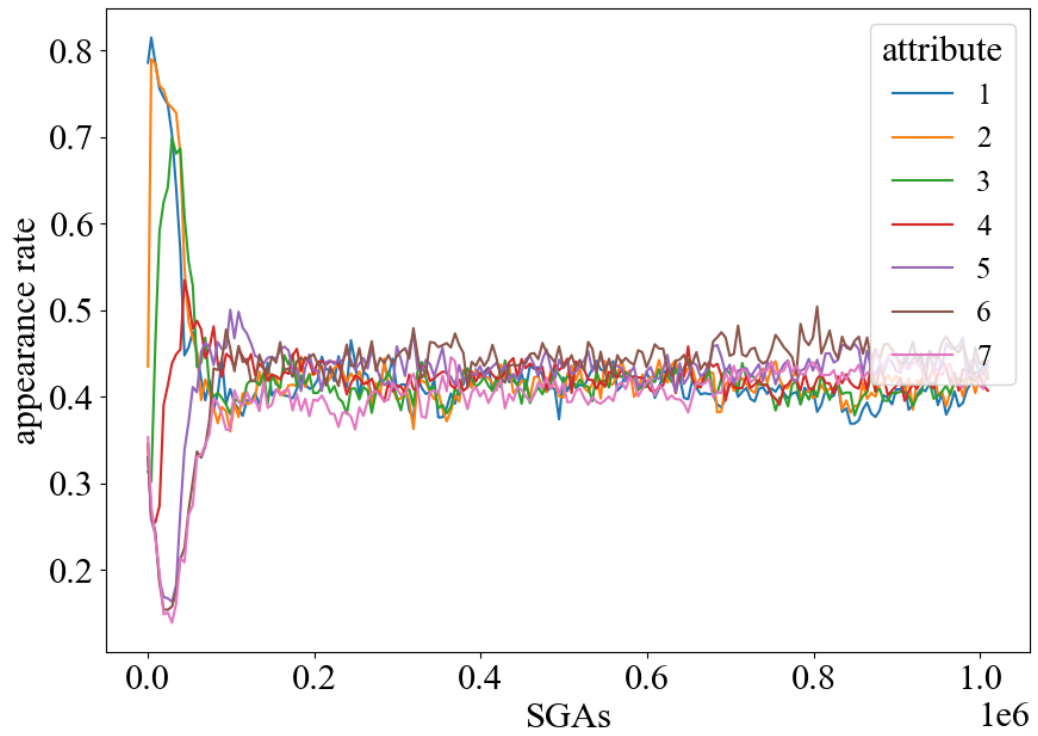

**Fig. 7.** Median appearance rate of attributes for all agents over SGAs of training, as selected by Version 2.

Finally, regarding the evaluation of the GA based PCG performance, the generated SG content in terms of frequency of presented attributes in NPC profiles is balanced in overall (Fig. 7). However, as displayed in Fig. 6 at any given time only 3 PCG generated attributes are dominant. This is the main difference of versions 2 and 3 against version 1, where all attributes appear with the same frequency constantly, due to the random nature of their selection. This is an indication of achievement of the PCG's desired goal of providing SG content gradually, in accordance with the agent's learning process. When the agents exhibit a winning pattern against certain attributes, versions 2 and 3 rotate them towards selections of attributes the agent is not yet familiar with. When they exhibit a losing pattern, attributes are rotated in a manner which does not increase difficulty abruptly and facilitate learning. This is further substantiated by the correspondence between changes in the agents' win rate and rotations in attributes shown in Fig. 6. This observation is in line with the GA's directive to adapt content to human players based on their interaction with the SG and provides further insight about possible similarities between DRL GT agent and human player behavior. The capacity of the proposed framework to produce data which provide this type of insights regarding procedurally generated content highlights its potential value.

In terms of the employed DRL agents, a readily available and well-established solution was selected, PPO, which, despite its versatility and overall capabilities, is not specifically developed for GT. Expanding on future studies to include and compare between a broader variety of DRL models and algorithms could yield invaluable insights on the performance of different agents and provide improved robustness to the findings. Furthermore, the showcase experiment presented in this article featured different versions of one existing SG, thus limiting insight gained regarding its generalization capabilities. Additionally, more refined approaches as baselines for comparison (e.g. structured randomness) could help improve fairness and better highlight the specific advantages of the studied PCG methods. Future work can focus on more showcase experiments, featuring different SGs from various genres. Those games could help broaden the reach of this study, by incorporating different types of PCG, and justify the framework's generalization capabilities. The addition of qualitative interviews of human participants facing the PCG techniques could provide added value for interpreting the framework's results. Another future direction could include the addition of an automated interpreter to the proposed framework, limiting the need for human interpretation of the produced metrics and potentially increasing its modularity and scalability.

Finally, the presented work provides initial insight regarding the interpretability and explainability of DRL agents' actions and decision-making when reacting to procedurally generated content. Obtaining this insight from the agents' behavior is important, especially for SGs in critical domains such as healthcare, towards uncovering potential SG design flaws and ensuring the fulfillment of their serious educational or healthcare purposes. The incorporation of XAI techniques and models in the proposed framework could reveal how DRL agents align their decisions and actions with the serious purpose of the game and fulfill the intended learning outcomes.

## VII. Conclusion

In this paper, a flexible and modular methodology for automated evaluation of PCG integration in SGs through DRL agents is presented. To validate the proposed framework, three different versions of a SG have been employed, one incorporating random content generation and two GA based PCG. The obtained results advocate towards the ability of the evaluation framework to discern between differences in effectiveness of different PCG versions and produce meaningful insight.

## References

[1] J. Togelius, E. Kastbjerg, D. Schedl, and G. N. Yannakakis, "What is procedural content generation? Mario on the borderline," in *Proceedings of the 2nd International Workshop on Procedural Content Generation in Games*, in PCGames '11. New York: Association for Computing Machinery, Jun. 2011, pp. 1–6. doi: 10.1145/2000919.2000922.

[2] G. N. Yannakakis and J. Togelius, "Experience-Driven Procedural Content Generation," *IEEE Trans. Affect. Comput.*, vol. 2, no. 3, pp. 147–161, Jul. 2011, doi: 10.1109/T-AFFC.2011.6.

[3] W. L. Raffe, F. Zambetta, X. Li, and K. O. Stanley, "Integrated Approach to Personalized Procedural Map Generation Using Evolutionary Algorithms," *IEEE Trans. Comput. Intell. AI Games*, vol. 7, no. 2, pp. 139–155, Jun. 2015, doi: 10.1109/TCIAIG.2014.2341665.

[4] D. Djaouti, J. Alvarez, and J.-P. Jessel, "Classifying serious games: the G/P/S model," in *Handbook of research on improving learning and motivation through educational games: Multidisciplinary approaches*, IGI global, 2011, pp. 118–136.

[5] C. Schrader, J. Brich, J. Frommel, V. Riemer, and K. Rogers, "Rising to the Challenge: An Emotion-Driven Approach Toward Adaptive Serious Games," in *Serious Games and Edutainment Applications : Volume II*, M. Ma and A. Oikonomou, Eds., Cham: Springer International Publishing, 2017, pp. 3–28. doi: 10.1007/978-3-319-51645-5_1.

[6] B. Said, L. Cheniti-Belcadhi, and G. El Khayat, "An Ontology for Personalization in Serious Games for Assessment," in *2019 IEEE Second International Conference on Artificial Intelligence and Knowledge Engineering (AIKE)*, Jun. 2019, pp. 148–154. doi: 10.1109/AIKE.2019.00035.

[7] G. N. Yannakakis, "Siren: Towards adaptive serious games for teaching conflict resolution," in *Proceedings of the 4th European Conference on*

*Games Based Learning*, Copenhagen, Denmark: Academic Publishing Limited, Oct. 2010, pp. 412–417.

[8] A. Moradi-Karkaj, “Serious Interactive Digital Narrative: Explorations in Personalization and Player Experience Enrichment,” in *2021 International Serious Games Symposium (ISGS)*, Nov. 2021, pp. 35–42. doi: 10.1109/ISGS54702.2021.9685015.

[9] H. Bomström, M. Kelanti, J. Lappalainen, E. Annanperä, and K. Liukkunen, “Synchronizing Game and AI Design in PCG-Based Game Prototypes,” in *International Conference on the Foundations of Digital Games*, Bugibba Malta: ACM, Sep. 2020, pp. 1–8. doi: 10.1145/3402942.3402989.

[10] C. P. Schultz and R. D. Bryant, *Game Testing: All in One*. Mercury Learning and Information, 2016.

[11] J. Bergdahl, C. Gordillo, K. Tollmar, and L. Gisslén, “Augmenting Automated Game Testing with Deep Reinforcement Learning,” in *2020 IEEE Conference on Games (CoG)*, Aug. 2020, pp. 600–603. doi: 10.1109/CoG47356.2020.9231552.

[12] J. Liu, S. Snodgrass, A. Khalifa, S. Risi, G. N. Yannakakis, and J. Togelius, “Deep learning for procedural content generation,” *Neural Comput. Appl.*, vol. 33, no. 1, pp. 19–37, Jan. 2021, doi: 10.1007/s00521-020-05383-8.

[13] S. F. Gudmundsson *et al.*, “Human-Like Playtesting with Deep Learning,” in *2018 IEEE Conference on Computational Intelligence and Games (CIG)*, Aug. 2018, pp. 1–8. doi: 10.1109/CIG.2018.8490442.

[14] Y. Matsuo *et al.*, “Deep learning, reinforcement learning, and world models,” *Neural Netw.*, vol. 152, pp. 267–275, Aug. 2022, doi: 10.1016/j.neunet.2022.03.037.

[15] A. Plaat, *Deep Reinforcement Learning, a textbook*. 2022. doi: 10.1007/978-981-19-0638-1.

[16] J. Schulman, F. Wolski, P. Dhariwal, A. Radford, and O. Klimov, “Proximal Policy Optimization Algorithms,” Aug. 28, 2017, *arXiv*: arXiv:1707.06347. doi: 10.48550/arXiv.1707.06347.

[17] A. Sehgal, N. Ward, H. La, and S. Louis, “Automatic Parameter Optimization Using Genetic Algorithm in Deep Reinforcement Learning for Robotic Manipulation Tasks,” Nov. 01, 2022, *arXiv*: arXiv:2204.03656. doi: 10.48550/arXiv.2204.03656.

[18] V. Mnih *et al.*, “Human-level control through deep reinforcement learning,” *Nature*, vol. 518, no. 7540, Art. no. 7540, Feb. 2015, doi: 10.1038/nature14236.

[19] S. (Sihang) Guo *et al.*, “Machine versus Human Attention in Deep Reinforcement Learning Tasks,” in *Advances in Neural Information Processing Systems*, Curran Associates, Inc., 2021, pp. 25370–25385. Accessed: Jan. 12, 2023. [Online]. Available: https://proceedings.neurips.cc/paper/2021/hash/d58e2f077670f4de9cd7963c857f2534-Abstract.html

[20] G. Liu *et al.*, “Inspector: Pixel-Based Automated Game Testing via Exploration, Detection, and Investigation,” in *2022 IEEE Conference on Games (CoG)*, Aug. 2022, pp. 237–244. doi: 10.1109/CoG51982.2022.9893630.

[21] S. Roohi, A. Relas, J. Takatalo, H. Heiskanen, and P. Hämäläinen, “Predicting Game Difficulty and Churn Without Players,” in *Proceedings of the Annual Symposium on Computer-Human Interaction in Play*, in CHI PLAY ’20. New York, NY, USA: Association for Computing Machinery, Nov. 2020, pp. 585–593. doi: 10.1145/3410404.3414235.

[22] A. Seyderhelm and K. Blackmore, “Systematic Review of Dynamic Difficulty Adaption for Serious Games: The Importance of Diverse Approaches,” *SSRN Electron. J.*, Jan. 2021, doi: 10.2139/ssrn.3982971.

[23] Y. Zheng *et al.*, “Wuji: Automatic Online Combat Game Testing Using Evolutionary Deep Reinforcement Learning,” in *2019 34th IEEE/ACM International Conference on Automated Software Engineering (ASE)*, Nov. 2019, pp. 772–784. doi: 10.1109/ASE.2019.00077.

[24] R. Fuchs, R. Gieseke, and A. Dockhorn, “Personalized Dynamic Difficulty Adjustment Imitation Learning Meets Reinforcement Learning,” in *2024 IEEE Conference on Games (CoG)*, Aug. 2024, pp. 1–2. doi: 10.1109/CoG60054.2024.10645659.

[25] S. Dahlskog, B. Horn, N. Shaker, G. Smith, and J. Togelius, “A Comparative Evaluation of Procedural Level Generators in the Mario AI Framework,” Apr. 2014.

[26] A. Summerville, “Expanding Expressive Range: Evaluation Methodologies for Procedural Content Generation,” *Proc. AAAI Conf. Artif. Intell. Interact. Digit. Entertain.*, vol. 14, no. 1, Art. no. 1, Sep. 2018, doi: 10.1609/aiide.v14i1.13012.

[27] V. Volz, B. Naujoks, P. Kerschke, and T. Tušar, “Tools for Landscape Analysis of Optimisation Problems in Procedural Content Generation for Games,” *Appl. Soft Comput.*, vol. 136, p. 110121, Mar. 2023, doi: 10.1016/j.asoc.2023.110121.

[28] S. Roohi, J. Takatalo, C. Guckelsberger, and P. Hämäläinen, “Review of Intrinsic Motivation in Simulation-based Game Testing,” in *Proceedings of the 2018 CHI Conference on Human Factors in Computing Systems*, in CHI ’18. New York, NY, USA: Association for Computing Machinery, Apr. 2018, pp. 1–13. doi: 10.1145/3173574.3173921.

[29] C. Holmgård, M. C. Green, A. Liapis, and J. Togelius, “Automated Playtesting With Procedural Personas Through MCTS With Evolved Heuristics,” *IEEE Trans. Games*, vol. 11, no. 4, pp. 352–362, Dec. 2019, doi: 10.1109/TG.2018.2808198.

[30] P. L. P. de Woillemont, R. Labory, and V. Corruble, “Automated Play-Testing through RL Based Human-Like Play-Styles Generation,” *Proc. AAAI Conf. Artif. Intell. Interact. Digit. Entertain.*, vol. 18, no. 1, Art. no. 1, Oct. 2022, doi: 10.1609/aiide.v18i1.21958.

[31] D. Hooshyar, M. Yousefi, M. Wang, and H. Lim, “A data-driven procedural-content-generation approach for educational games,” *J. Comput. Assist. Learn.*, vol. 34, no. 6, pp. 731–739, 2018, doi: 10.1111/jcal.12280.

[32] V. Volz *et al.*, “Capturing Local and Global Patterns in Procedural Content Generation via Machine Learning,” in *2020 IEEE Conference on Games (CoG)*, Aug. 2020, pp. 399–406. doi: 10.1109/CoG47356.2020.9231944.

[33] D. Perez-Liebana, S. Samothrakis, J. Togelius, T. Schaul, and S. Lucas, “General Video Game AI: Competition, Challenges and Opportunities,” *Proc. AAAI Conf. Artif. Intell.*, vol. 30, no. 1, Art. no. 1, Mar. 2016, doi: 10.1609/aaai.v30i1.9869.

[34] D. Perez-Liebana, J. Liu, A. Khalifa, R. D. Gaina, J. Togelius, and S. M. Lucas, “General Video Game AI: A Multitrack Framework for Evaluating Agents, Games, and Content Generation Algorithms,” *IEEE Trans. Games*, vol. 11, no. 3, pp. 195–214, Sep. 2019, doi: 10.1109/TG.2019.2901021.

[35] C. Guerrero-Romero and D. Perez-Liebana, “MAP-Elites to Generate a Team of Agents that Elicits Diverse Automated Gameplay,” in *2021 IEEE Conference on Games (CoG)*, Aug. 2021, pp. 1–8. doi: 10.1109/CoG52621.2021.9619142.

[36] C. Guerrero-Romero, S. Lucas, and D. Perez-Liebana, “Beyond Playing to Win: Creating a Team of Agents With Distinct Behaviors for Automated Gameplay,” *IEEE Trans. Games*, vol. 15, no. 3, pp. 469–482, Sep. 2023, doi: 10.1109/TG.2023.3241864.

[37] S. Stahlke, A. Nova, and P. Mirza-Babaei, “Artificial Players in the Design Process: Developing an Automated Testing Tool for Game Level and World Design,” in *Proceedings of the Annual Symposium on Computer-Human Interaction in Play*, Virtual Event Canada: ACM, Nov. 2020, pp. 267–280. doi: 10.1145/3410404.3414249.

[38] C. Paduraru, M. Paduraru, and A. Stefanescu, “Automated game testing using computer vision methods,” in *2021 36th IEEE/ACM International Conference on Automated Software Engineering Workshops (ASEW)*, Nov. 2021, pp. 65–72. doi: 10.1109/ASEW52652.2021.00024.

[39] M. Genesereth and M. Thielscher, “Game Description,” in *General Game Playing*, M. Genesereth and M. Thielscher, Eds., in Synthesis Lectures on Artificial Intelligence and Machine Learning. , Cham: Springer International Publishing, 2014, pp. 13–29. doi: 10.1007/978-3-031-01569-4_2.

[40] K. Mitsis, E. Kalafatis, K. Zarkogianni, G. Mourkousis, and K. S. Nikita, “Procedural content generation based on a genetic algorithm in a serious game for obstructive sleep apnea,” in *2020 IEEE Conference on Games (CoG)*, Aug. 2020, pp. 694–697. doi: 10.1109/CoG47356.2020.9231785.

[41] M. Towers *et al.*, “Gymnasium: A Standard Interface for Reinforcement Learning Environments,” Nov. 08, 2024, *arXiv*: arXiv:2407.17032. doi: 10.48550/arXiv.2407.17032.

[42] *DLR-RM/stable-baselines3*. (Dec. 03, 2024). Python. DLR-RM. Accessed: Dec. 03, 2024. [Online]. Available: https://github.com/DLR-RM/stable-baselines3

[43] A. Raffin, A. Hill, A. Gleave, A. Kanervisto, M. Ernestus, and N. Dormann, “Stable-baselines3: reliable reinforcement learning implementations,” *J. Mach. Learn. Res.*, vol. 22, no. 1, p. 268:12348-268:12355, Jan. 2021.

[44] D. Kristan, P. Bessa, R. Costa, and C. Vaz de Carvalho, “Creating Competitive Opponents for Serious Games through Dynamic Difficulty Adjustment,” *Information*, vol. 11, no. 3, Art. no. 3, Mar. 2020, doi: 10.3390/info11030156.